\def\year{2022}\relax
\documentclass[letterpaper]{article} % DO NOT CHANGE THIS
\usepackage{aaai22}  % DO NOT CHANGE THIS
\usepackage{times}  % DO NOT CHANGE THIS
\usepackage{helvet}  % DO NOT CHANGE THIS
\usepackage{courier}  % DO NOT CHANGE THIS
\usepackage[hyphens]{url}  % DO NOT CHANGE THIS
\usepackage{graphicx} % DO NOT CHANGE THIS
\usepackage{natbib}  % DO NOT CHANGE THIS AND DO NOT ADD ANY OPTIONS TO IT
\usepackage{caption} % DO NOT CHANGE THIS AND DO NOT ADD ANY OPTIONS TO IT
\DeclareCaptionStyle{ruled}{labelfont=normalfont,labelsep=colon,strut=off} % DO NOT CHANGE THIS
\usepackage{amsfonts}
\usepackage{amsmath}
\usepackage{multirow}
\usepackage{amssymb}% http://ctan.org/pkg/amssymb
\usepackage{pifont}% http://ctan.org/pkg/pifont
\usepackage{bbold}
\usepackage{xcolor}

\usepackage{amsmath, bm}
\usepackage{amsfonts,amssymb}

\usepackage{algorithm}
\usepackage{algorithmic}

\usepackage{newfloat}
\usepackage{listings}
\floatstyle{ruled}
\newfloat{listing}{tb}{lst}{}
\floatname{listing}{Listing}

\title{Single-domain Generalization in Medical Image Segmentation via \\ Test-time Adaptation from Shape Dictionary}
\author {
        Quande Liu\textsuperscript{\rm 1},
        Cheng Chen\textsuperscript{\rm 1},
        Qi Dou\textsuperscript{\rm 1},
        Pheng-Ann Heng\textsuperscript{\rm 1,2}\\
}
\affiliations {
    \textsuperscript{\rm 1} Department of Computer Science and Engineering, The Chinese University of Hong Kong \\
    \textsuperscript{\rm 2} Guangdong-Hong Kong-Macao Joint Laboratory of Human-Machine Intelligence-Synergy Systems, \\Shenzhen Institute of Advanced Technology, Chinese Academy of Sciences \\
}

\usepackage{bibentry}
\begin{document}
% \linenumbers
\maketitle

\begin{abstract}
Domain generalization typically  requires data from multiple source domains for model learning.
However, such strong assumption may not always hold in practice, especially in medical field where the data sharing is highly concerned and sometimes prohibitive due to privacy issue. 
This paper studies the important yet challenging single domain generalization problem, in which a model is learned under the worst-case scenario with only one source domain to directly generalize to different unseen target domains.
We present a novel approach to address this problem in medical image segmentation, which extracts and integrates the semantic shape prior information of segmentation that are invariant across domains and  can be well-captured even from single domain data to facilitate segmentation under distribution shifts. Besides, a test-time adaptation strategy with dual-consistency regularization is further devised to promote dynamic incorporation of these shape priors under each unseen domain to improve model generalizability. 
Extensive experiments on two medical image segmentation tasks demonstrate the consistent improvements of our method across various unseen domains, as well as its superiority over state-of-the-art approaches in addressing domain generalization under the worst-case scenario.

\end{abstract}

\section{Introduction}

Deep networks are notoriously difficult to generalize to new domains due to distribution shift between training and testing data acquired at different situations.
Domain generalization (DG) has recently emerged to improve the generalizability of deep networks on unseen domains.
Most existing DG methods typically require to learn from multi-source distributions to extract representations that are robust to distribution shifts~\cite{li2017deeper}. 
However, the prerequisite to aggregate multi-domain data may not be always achievable in practice. 
In the field of medical imaging, for example, the data present privacy concerns and are prohibitive to be shared before taking complex procedures including ethical approval and patient consent~\cite{price2019privacy}.
In this regard, directly generalizing a deep model trained with only single-domain data to any unseen domain is increasingly important and has wider applicability in medical applications.

This problem is \textit{single domain generalization} (SDG), the worst-case scenario of DG with only one source domain available for model learning. 
Under this constraint, previous DG methods depending heavily on accessing multi-domain data can hardly work well or lack feasibility in the SDG setting. For instance, the adversarial learning method~\cite{li2018domain} learns domain-invariant representations by regularizing the feature space of multiple data sources, which is infeasible in the case when only one domain is provided for training.
Recently, a few studies have started to address SDG problem with the typical solution of adversarial data augmentation ~\cite{volpi2018generalizing,qiao2020learning,li2021progressive}, which aims to synthesize fictitious samples from single-domain data to potentially simulate the data distributions of unseen domains. 
\textcolor{black}{However, since it is extremely difficult to anticipate the distribution of test data, these methods are less effective in medical imaging where the domain discrepancies could be large due to complex factors and hardly be simulated from source data (e.g., the domain shift caused by whether using endorectal coil in prostate MRI scanning).}
Moreover, medical data are usually high-dimensional with tremendous tissue details, making it non-trivial to create high-quality virtual data even after careful model tuning.

Regarding these limitations, we instead aim to address the challenging SDG problem in a simple yet effective way, which is inspired by two key observations in medical image segmentation tasks. 
First, though the image appearances differ across clinical centers, the semantic shape of segmentation is highly consistent across data sources reflecting the general anatomical structures in medical images.
Encouragingly, such shape information can be well-represented even within one single domain, without necessarily aggregating data from multiple clinical sources.
Second, though the target distributions of large discrepancies can hardly be simulated during model training, they can instead be explored at test time by considering the given inference sample  as a hint.
This enables us to mitigate the image distribution bias of single-domain training by adjusting model parameters for each unseen data distribution accordingly at test time. 
% \clearpage

Based on these insights, we present a new method for addressing SDG named \textit{Test-time Adaptation from Shape Dictionary} (TASD). 
The core idea is to integrate the general semantic shapes extracted from single-domain data into the segmentation network, and further effectively leverage these prior information through test-time adaptation to generalize the single-domain model to any unseen domains.
To this end, our TASD first establishes a dictionary learning strategy to extract a set of explicit shape priors from the single source domain. 
A regression branch is then embedded onto the network, which produces weighting coefficients to integrate these dictionary items, so that the segmentation could be derived by jointly recognizing the sample features as well as the pre-collected shape priors to facilitate generalization. Next, when deploying the model to new domains, we have devised a dual-consistency regularization mechanism to adapt the model parameters at test time, which jointly enforces the consistency of shape coefficients and segmentation predictions under different input perturbations applied to the inference data. Such adaptation strategy allows to dynamically update the model parameters especially for the regression branch at test time, which enables the model to adaptively utilize the shape prior information under any unseen data distributions to improve model generalizability.

Our main contributions are highlighted as follows: 
\begin{itemize}
\item We present a novel approach to address the challenging single domain generalization problem for medical image segmentation, by explicitly exploiting the general semantic shape priors that are extractable from single-domain data and are generalizable across domains to assist domain generalization under the worst-case scenario.

\item We also devise an effective test-time adaptation strategy with dual-consistency regularization, which allows to adaptively utilize the shape prior information at any unseen data distributions to improve model generalizability.

\item We conduct extensive experiments on two typical medical image segmentation tasks, i.e., prostate MRI and retinal fundus image segmentation. 
With only single-domain training, our method significantly improves generalization performance on many unseen domains, outperforming the state-of-the-arts. 

\end{itemize}

\section{Related Work}

\textbf{Domain generalization} aims to generalize models to completely unseen domains without accessing target data during training~\cite{chattopadhyay2020learning,du2020learning,hoffer2020augment,liu2021feddg,seo2019learning,yue2019domain,fan2021adversarially,dou2021federated}. 
Previous progress on DG is mainly made by enabling models to learn from multi-source domain data to reduce the model bias. 
A plenty of methods aim to extract domain-invariant representations by aligning the feature space of multiple data sources~\cite{muandet2013domain,ghifary2015domain,hsu2017learning,motiian2017unified,li2018domain,li2018deep}. 
Some other studies have introduced self-supervised learning~\cite{huang2020self} or model-agnostic meta-learning~\cite{li2017learning,dou2019domain,balaji2018metareg,li2019feature} to regularize the model training with multiple domains to learn generalizable representations.
Recently, the worst-case scenario with only one source domain available for model training has been attempted in
\cite{qiao2020learning,volpi2018generalizing,wang2021learning,li2021progressive} with adversarial domain augmentation.
Typically, ~\cite{qiao2020learning} propose a method which incorporates meta-learning and Wasserstein auto-encoder to facilitate the learning efficiency from generated data. Considering the large domain discrepancies and the difficulty in data generation for medical images,
we instead tackle the challenging SDG problem from a new perspective by leveraging the domain-invariant shape information of medical anatomies and promoting effective test-time utilization of these learned shape priors. 

\textbf{Dictionary learning} seeks to find a set of basic elements to compose a dictionary such that a given input can be well represented by a sparse linear combination of these learned elements. Successful applications of dictionary learning have emerged in various image recognition tasks~\cite{zhang2012towards,heimann2009statistical}.
For example, dictionary learning is employed in \cite{vu2017fast} to learn a set of common patterns and class-specific patterns from different objects for image classification.
~\cite{cao2015sled} present semantic label dictionary which explicitly fuses the label information into dictionary representation and explores the semantic correlations between co-occurrence labels for multi-label image annotation. 
In medical applications, dictionary learning is conducted in \cite{tong2015discriminative} to find a set of most representative atlases, which are then applied for label propagation to solve a multi-organ segmentation task.  
In our work, we employ dictionary learning to explicitly extract the shape priors of medical anatomies from single-domain data, and more importantly, we integrate the learned shape dictionary into deep models to facilitate the generalization on unseen domains. To our best knowledge, this is the first work to leverage dictionary learning to solve model generalization problem.

\textbf{Test-time learning} is recently proposed to utilize the distribution information from the target data presented at test time to quickly adapt models with a few gradient steps~\cite{chen2021source,Pandey_2021_CVPR}.
The main differences of prior works lie in how to devise the objective for driving test-time learning and which part of network parameters to be updated at test time. 
For instance, the pioneer work~\cite{sun2020test} exploits the test data via an auxiliary branch with self-supervision to adapt the encoder of a classification model.
Later on, Wang et al.~\shortcite{wang2020fully} penalize the entropy of model predictions on test data to adapt the batch normalization layer for generalization purpose. 
For test-time adaptation on medical image segmentation, Karani et al.~\shortcite{karani2020test} introduce a denoising autoencoder into test-time learning to adapt an image normalization module. This method requires iterative refinement of model predictions and takes nearly 12 minutes for the adaptation on one medical volume data, which therefore cannot satisfy the real-time adaptation requirement in clinical scenarios. 
Compared with these methods, our test-time update is tightly-coupled with the learned generalizable shape dictionary, and is driven by the lightweight dual-consistency regularization mechanism which enables the model to dynamically utilize the shape priors according to various data distributions at test time.

\section{Methodology}
\begin{figure*}[t]
	\centering
	\includegraphics[width=\textwidth]{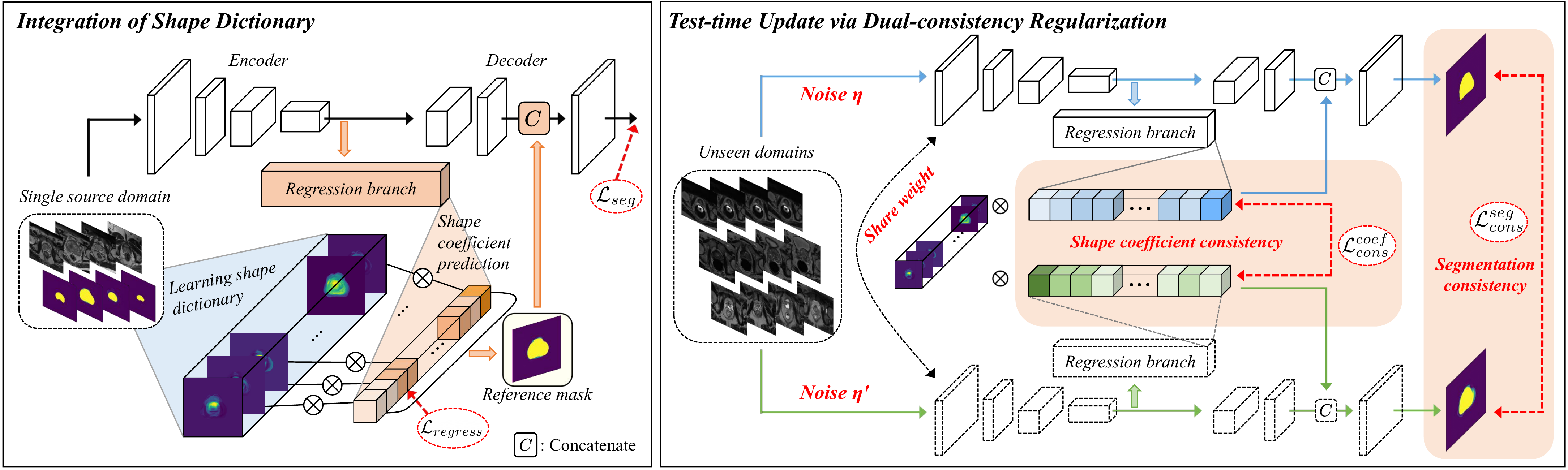}
	\caption{Overview of our proposed \textit{Test-time Adaptation from Shape Dictionary} (TASD). During training, we establish dictionary learning to extract the explicit shape priors from single domain data and
	integrate these prior information into the network by introducing a regression branch to produce shape coefficients for combining these dictionary items. When testing on new domains, we dynamically adapt the model parameters especially on the regression branch to adaptively utilize the prior information to improve generalization, which is driven by dual-consistency regularization mechanism which jointly regularizes the consistency of shape coefficients and segmentation predictions under different perturbations ($\eta, \eta'$) onto the inference data.}
	\label{fig:method}
\end{figure*}

An overview of our method \textit{Test-time Adaptation from Shape Dictionary} is shown in Fig.~\ref{fig:method}. This section describes the two main components of TASD followed by its technical details of source domain training and test-time update. 

% \clearpage
\subsection{Integrating Shape Dictionary for SDG}
\textbf{Construction of shape dictionary:} The semantic shapes of segmentation reflect the underlying anatomical structures in medical images, thus are independent of the scanner effects from observed domains. In light of this, we expect to harness this general shape information, which can be well-captured even within single source domain,  to boost model generalization in SDG. Motivated by the nature of dictionary learning to learn representative elements, we employ it to explicitly extract the representative shape priors from single-domain data, and further effectively integrate the shape information into the task model to provide references for the segmentation at unseen domains.

Specifically, dictionary learning is to find a set of basic elements (called dictionary), such that a given input can be approximated as their sparse linear combinations. In our case, denote $\mathbf{S} = \{(\mathbf{x}_i, \mathbf{y}_i)\}_{i=1}^{N}$ as the single source domain data containing $N$ pairs of data and labels. We intend to find a dictionary $\mathbf{D}$ consisting of $K$ explicit shape templates, i.e., $\mathbf{D}=\{\mathbf{d}_1, \mathbf{d}_2, \dots, \mathbf{d}_K\}$, that is able to represent the diverse segmentation masks $\{\mathbf{y}_i\}_{i=1}^{N}$ in the task. 
To ensure that the shape dictionary learned from one domain is representative enough to be extended to unseen domains, two principles need to be considered for its construction: (1) $K \ll N$, the number of shape templates should be much smaller than the number of masks in the single source data (e.g., $K = 48$ for $N= 400$), to reduce overfitting of the learned dictionary, (2) the mask $\mathbf{y}_i$ should be represented with as few elements as possible (i.e., the coefficients for linear combination should be sparse), to maximize the representation power of each template in the shape dictionary. 
For simplicity, we consider a binary segmentation task (i.e., $\mathbf{y}_i\in\mathbb{R}^{H \times W}$), while the extension to multi-class segmentation is straightforward by constructing a shape dictionary for each class in the same way. Then the generation of dictionary is equivalent to solving the optimization problem with respect to the shape dictionary $\mathbf{D}\in \mathbb{R}^{K\times H \times W}$ and the coefficients $\bm{\alpha}=[\bm{\alpha}_1, \bm{\alpha}_2, \dots, \bm{\alpha}_N]$ ($\bm{\alpha}_i\in\mathbb{R}^K$ denotes the learned coefficients of dictionary for each $\mathbf{y}_i$) as:
\begin{equation}
\begin{aligned}
    & \mathop{\arg\min}_{\mathbf{D}, \bm{\alpha}}
    && \sum_{i=1}^N (||\mathbf{y}_i - \sum_{j=1}^K \mathbf{d}_j \bm{\alpha}_{ij}||_2^2 + \lambda ||\bm{\alpha}_i||_1), \\
    % & \text{subject to} && ||\mathbf{D}||\\
\end{aligned}
\label{eq:dictionary_generation}
\end{equation}
where $\lambda$ is a balancing parameter to regularize the learned coefficients. In our implementation, we fixed $\lambda$ as $1.2/\sqrt{m}$ (m denotes the input resolution, i.e., $H\times W$) as this is a general solution in dictionary learning to yield sparse coefficients. The optimization problem in Eq.~\ref{eq:dictionary_generation} can be solved using the well-established online dictionary learning algorithm~\cite{mairal2009online}, which iteratively draws each image mask $\mathbf{y}_i$ to find its sparse coefficients with least-angle regression and update the dictionary $\mathbf{D}$ with block-coordinate descent. Meanwhile, this learning algorithm is parameter-free hence does not need parameter-tuning. We show in experiments that the shape dictionary learned in this way is representative and can well generalize to unseen domains.
\\\textbf{Integration of shape dictionary into deep network:} After obtaining the learned shape dictionary, we further incorporate it into the deep model, aiming to enable the segmentation to be performed with references from these shape priors for boosting generalization. To this end, we embed a regression branch onto the network (see the left part of Fig.~\ref{fig:method}), which takes the encoded features of a given sample as input, and output the shape coefficients $\bm{\hat{\alpha}}$ for combining the shape priors according to the sample features. Specifically, the pre-extracted shape priors in $\mathbf{D}$ are linearly combined according to the shape coefficients $\bm{\hat{\alpha}}$ to generate a reference mask:
\begin{equation}
\begin{aligned}
\bm{M} = \sum_{j=1}^K \mathbf{d}_j  \bm{\hat{\alpha}}_{ij}
\end{aligned}
\label{eq:reference_mask}
\end{equation}
which is then concatenated with network features before the last convolutional block of decoder for further refinement and generate the final segmentation mask. We did not choose to insert the reference mask at early layers since this requires downsampling the reference mask, which would destroy the shape prior information to some extent. With the help of the embedded regression branch, the model predictions can be derived by jointly exploring the  sample features as well as the shape priors extracted from single-domain data to improve model generalization.

To provide direct supervision for the regression branch to improve the quality of reference mask, we fetch the coefficients $\bm{\alpha}$ optimized from Eq.~\ref{eq:dictionary_generation} (obtained during dictionary learning) as the ground truth of $\bm{\hat{\alpha}}$. Since $\bm{\alpha}$ is regularized to be sparse during dictionary learning, we adopt the cosine similarity to minimize the distance between $\bm{\hat{\alpha}}$ and $\bm{\alpha}$ instead of the L$1$ or L$2$ loss, which is expressed as:
\begin{equation}
\begin{aligned}
    \mathcal{L}_{{regress}} (\mathbf{x}_i) = 1- \frac{\bm{\hat{\alpha}}_i \cdot \bm{\alpha}_i}{\text{max} (||\bm{\hat{\alpha}}_i||_2\cdot||\bm{\alpha}_i||_2, \epsilon)},
\end{aligned}
\end{equation}
where $\epsilon$ is a small value set as 1e-8 to avoid division by zero. 
\subsection{Dual-consistency Regularized Test-time Update}

With embedding an regression branch onto the network, the shape priors have been integrated to assist the segmentation problem. However, a potential issue is that the model's ability to utilize these prior information (i.e., parameters of the regression branch and original network) are learned under the narrow single-domain data distributions.  When generalizing to unseen domains with distribution shifts, the reliability of generated shape coefficients cannot be guaranteed, which would limit the shape priors to be effectively utilized. Given the observation that the inference sample presented at test time can give us a hint about its distribution information, we expect to dynamically adapt the model parameters at test time by exploring this hint, so that the shape priors can be effectively utilized under any unseen data distributions to improve generalization performance.

Crucially, the test-time update in our case should be tightly-coupled with the regression branch to ensure the correctness of shape coefficients under domain shifts. To this end, we formulate the adaptation as a consistency regularization mechanism, which can be flexibly imposed on both original network as well as the embedded regression branch. 
Specifically, as shown in the right part of Fig.~\ref{fig:method}, given an inference data $\mathbf{x}^t$ at test time, we first add different perturbations $\eta$ and $\eta'$ (e.g., adding Gaussian noise to input or dropout to network parameters) onto this sample, and then regularize the model to generate consistent predictions for the two perturbed samples to explore the inherent information of this testing data. Importantly, to fully adapt the parameters of regression branch apart from the original network, we impose the consistency regularization on both segmentation predictions and shape coefficients predictions. The intuition behind is that different input perturbations should not change the semantic anatomies contained in a medical image, hence the final segmentation mask of an inference sample as well as how our model leverages the pre-collected shape priors (i.e., shape coefficients) to compose the mask should not be changed. Formally, these two consistency regularization terms can be formulated as:
\begin{equation}
\begin{aligned}
    &\mathcal{L}_{cons}^{seg}(\mathbf{x}^t) = ||f(\mathbf{x}^t, \eta') - f(\mathbf{x}^t, \eta) ||_2^2, \\
    &\mathcal{L}_{cons}^{coef}(\mathbf{x}^t) =1- \frac{\bm{\hat{\alpha}}^t(\eta') \cdot \bm{\hat{\alpha}}^t(\eta)}{\text{max} (||\bm{\hat{\alpha}}^t(\eta')||_2\cdot||\bm{\hat{\alpha}}^t(\eta)||_2, \epsilon)},
    \label{equ:test^time_update}
\end{aligned}
\end{equation}
where $\mathcal{L}_{cons}^{seg}(\mathbf{x}^t)$ is the consistency regularization on segmentation results with mean square error loss, in which $f(\cdot)$ denotes the network, $\eta$ and $\eta'$ are different perturbations applied to the same testing data $\mathbf{x}^t$; and $\mathcal{L}_{cons}^{coef}(\mathbf{x}^t)$ denotes the consistency regularization on shape coefficients with cosine distance loss, in which $\bm{\hat{\alpha}}^t(\cdot)$ is the shape coefficient predictions for $\mathbf{x}^t$ under perturbation $(\cdot)$. Our dual-consistency regularization objective can then be expressed as:
\begin{equation}
\begin{aligned}
  &\mathcal{L}_{cons}^{dual}(\mathbf{x}^t) = \mathcal{L}_{cons}^{seg}(\mathbf{x}^t) + \mathcal{L}_{cons}^{coef}(\mathbf{x}^t),
    \label{equ:test^time_update}
\end{aligned}
\end{equation}

At test time, we only perform one step of gradient descent by optimizing Eq.~\ref{equ:test^time_update} to adapt the parameters of source model given each inference data. Based on our experiments, such adaptation strategy only requires in average 56ms to process a 2D medical image and 0.91s for a medical volume data, hence can be performed efficiently. Meanwhile, we only adapt the feature normalization parameters, i.e., the parameters of the batch normalization layers in the source model, which enables the model to adaptively normalize the features according to the data distributions of each unseen domain. This yields more stable results in our experiments compared with adapting all network parameters, which could be explained by that the deep network is overly parameterized 
hence adapting the whole model is prone to affect its original discriminability.  
It is worthy to note that the adaptation strategy in our method is performed in an online manner, i.e., when receiving test sample $\mathbf{x}^t_j$ at time point $j$, the model state is initialized with the parameters updated from the previous sample $\mathbf{x}^t_{j-1}$ instead of the original source model. This enables the model to fully utilize the distribution information from all previous samples to boost generalization, and is also consistent with the clinical scenario where testing samples usually arrive sequentially.

\subsection{Training Objectives and Strategies}
In our framework, we first learned the shape dictionary from single-domain data by solving Eq.~\ref{eq:dictionary_generation} with online dictionary learning algorithm~\cite{mairal2009online}. Then, we integrated the shape dictionary that contains the shape prior information into segmentation network, and the overall objective to train the network at single source domain is expressed as:
\begin{equation}
\begin{aligned}
    \mathcal{L}(\mathbf{x}_i) =
    \mathcal{L}_{seg}(\mathbf{x}_i) + \beta \mathcal{L}_{regress}(\mathbf{x}_i), 
\end{aligned}
\end{equation}
which is composed of the segmentation loss and the shape coefficient regression loss with a weighting parameter $\beta$. We empirically set $\beta$ as 5 to balance the scale of the two loss terms. Once the training is done, we can generalize the model to unseen domains by dynamically adapting the model parameters as described above.

\section{Experiment}
We extensively evaluate our approach on two typical medical image  segmentation tasks: 1) prostate segmentation on T2-weighted MRI, and 2) optic disc and cup segmentation on retinal fundus images. 
Results of comparison with state-of-the-arts and ablation study are presented in the following.
\subsection{Datasets and Evaluation Metrics}
~~\textbf{Prostate MRI segmentation:}
We collect prostate T2-weighted MRIs from six different clinical centers out of three public datasets, including NCI-ISBI13~\cite{isbi}, I2CVB~\cite{i2cvb} and PROMISE12~\cite{promise12} datasets. These data are uniformly resized to 384$\times$384 in axial plane, and each data volume is normalized to have zero mean and unit variance in intensity values. 
Due to the large variance on the slice thickness of data from different clinical sites, we employ a 2D network as backbone. 

\textbf{Fundus image segmentation:}
We employ retinal fundus images from four medical institutions out of three public datasets, including REFUGE~\cite{orlando2020refuge}, Drishti-GS~\cite{sivaswamy2015comprehensive} and RIM-ONE-r3~\cite{fumero2011rim} datasets. 
For this task, we follow the fundus image pre-processing pipeline of ~\cite{wang2020dofe} to first train a simple U-Net to detect and crop ROIs around optic disc with the size of 800$\times$800, which are then resize to 384$\times$384 as network inputs. 
Since the ROI size of 800$\times$800 is about twice of the optic disc region, the location deviation is relaxed and we found that all optic disc and cup regions are covered by the cropped ROI even under data distribution shifts.  We average the performance of optic disc and cup segmentation as the task performance. 

\begin{figure}[t]
	\centering
	\includegraphics[width=0.47\textwidth]{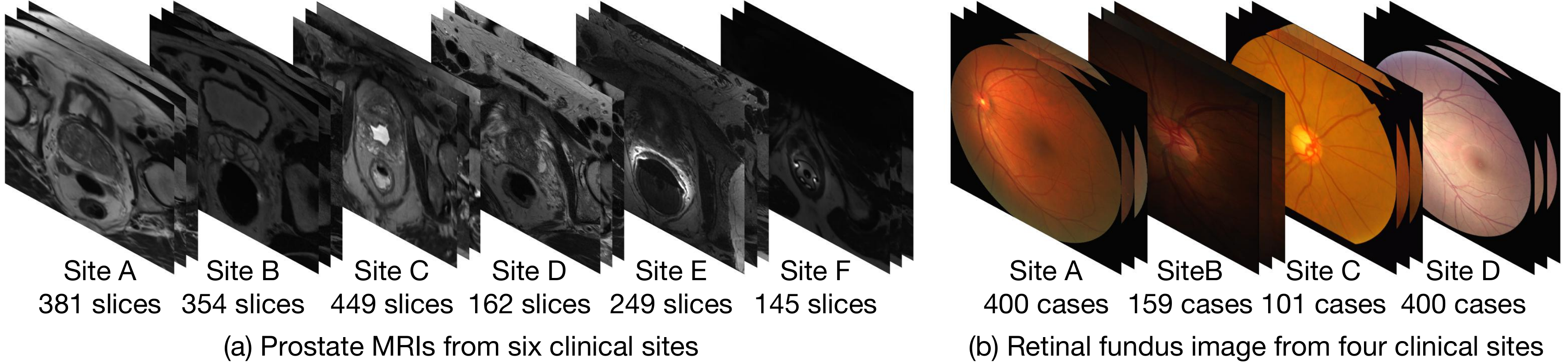}
    \caption{Representative cases and case number of each data source for prostate MRI and retinal fundus images.} 
 	\label{fig:dataset}
\end{figure}

In both tasks, data acquired from different clinical centers present discrepancies due to different imaging conditions. The representative cases and sample numbers of each dataset are shown in Fig.~\ref{fig:dataset}. 
For the evaluation, we employ two commonly-used metrics to quantitatively evaluate the segmentation results, including the Dice coefficient [\%] (the higher the better) and Hausdorff Distance (HD) [pixel] (the lower the better), which evaluate the segmentation results in terms of the whole object and surface region, respectively.

\subsection{Implementation Details}
In our implementation, the element number $K$ in the shape dictionary is set as 48, and we will investigate its effect in the ablation study. For input perturbations, two  dropout layers with dropout rate 0.5 are added before the bottleneck layer and the last convolution layer respectively, and Gaussian noise with magnitude 0.1 is added onto the input images. The step size for test-time update is 1e-3.
We employ an adapted Mix-residual-UNet~\cite{liu2020ms} as segmentation backbone.  
The standard data augmentation techniques are used to avoid overfitting, including random rotation and flipping (horizontal and vertical). The model is trained using Adam optimizer with momentum of 0.9 and 0.99, and the learning rate is initialized as 1e-3. We totally train 100 epochs on the single source domain as the network has converged, with batch size set as 5. 
The framework is implemented with Pytorch using one NVIDIA TitanXp GPU.

\subsection{Results of Single Domain Generalization}
\textbf{Experimental settings:} To fulfill SDG setting, for prostate MRI segmentation, we follow~\cite{karani2020test} to use a single domain dataset NCI-ISBI13 as the source domain (Site A) and take the other five data sources as different unseen target sites (Site B-F). 
For fundus image segmentation, we adopt the popular benchmark that is the training set of REFUGE as the single source domain (Site A), and the other three data sources as different target sites (Site B-D).

We compare our TASD with recent state-of-the-art methods in SDG and test-time learning, as well as popular DG methods that are not completely restricted to multiple source domains and can be applied in the SDG scenario, including:
\textbf{M-ADA (CVPR 2020)}~\cite{qiao2020learning}:
a SDG method with adversarial domain augmentation, which develops Wasserstein auto-encoder to create virtual data. \textbf{TTT (ICML 2020)} ~\cite{sun2020test} a test-time learning method for generalization under distribution shifts with self-supervision of rotation prediction. \textbf{TTST (MIA 2021)}~\cite{karani2020test} a test-time learning method for generalization in medical image segmentation with self-training from a denoising autoencoder. \textbf{Tent (ICLR 2021)}~\cite{wang2020fully}: 
a test-time adaptation method by minimizing the entropy of model predictions.
\textbf{BigAug (TMI 2020)}~\cite{zhang2020generalizing}: a DG method 
on medical image segmentation that adopts extensive data transformations to promote general representation learning. \textbf{JiGen (CVPR 2019)}~\cite{carlucci2019domain}: a DG method with self-supervised learning that learns general representations by solving jigsaw puzzles. In SDG scenario, the \textbf{Baseline} setting denotes learning a model on the single source domain without using any generalization technique, which is different from general DG scenario where the baseline model is obtained by training with multiple source datasets~\cite{liu2020shape,wang2020dofe}.

\begin{table*}[!tbp]
    \renewcommand\arraystretch{1.2}
    \centering
        \caption{\small{Quantitative comparison of single domain generalization results on prostate MRI segmentation between different methods.}}
        \resizebox{1.0\textwidth}{!}{%
        \setlength\tabcolsep{5.0pt}
        \scalebox{1.00}{
        \begin{tabular}{c|c  c  c  c c c||c  c  c  c c c}
    %       \hline
            \hline
            \hline
            Unseen Site  & B &  C & D & E &F &Avg.  & B &  C & D & E  &F &Avg.\\
            \hline
            \hline
            &\multicolumn{6}{c||}{\textbf{Dice Coefficient (Dice, mean$\pm$std)~$\uparrow$ }} &\multicolumn{6}{c}{\textbf{Hausdorff Distance (HD,  mean$\pm$std)~$\downarrow$}}\\ 
            \hline
            \hline
            Baseline &83.8$\pm$5.3 &73.3$\pm$11.1 &72.6$\pm$7.0 &65.5$\pm$29.5 &78.7$\pm$7.8 &74.8 &40.9$\pm$34.9 &59.0$\pm$29.4 &59.5$\pm$21.2 &61.2$\pm$42.6 &37.2$\pm$20.9 &51.6\\
            \hline
            M-ADA~\cite{qiao2020learning} &86.2$\pm$4.4 &74.7$\pm$9.1 &80.9$\pm$4.9 &69.7$\pm$12.2 &79.5$\pm$9.3 &78.2 &\textbf{19.1$\pm$21.1} &46.1$\pm$28.1 &53.9$\pm$19.3 &54.2$\pm$19.6 &31.9$\pm$26.7 &41.0\\
            TTT~\cite{sun2020test} &83.5$\pm$5.9 &73.1$\pm$17.5 &75.3$\pm$7.8 &67.5$\pm$11.1 &81.5$\pm$5.9 &76.2 &26.4$\pm$22.1 &55.4$\pm$22.3 &54.8$\pm$25.5 &53.0$\pm$22.2 &21.8$\pm$19.2 &42.3\\
            TTST~\cite{karani2020test} &86.0$\pm$3.7 &74.8$\pm$10.5 &81.0$\pm$3.9 &74.0$\pm$8.4 &80.9$\pm$9.2 & 79.3 &20.5$\pm$20.7 &47.5$\pm$28.1 &41.4$\pm$19.7 &51.4$\pm$26.1 &34.5$\pm$25.5 &39.1\\
            Tent~\cite{wang2020fully} &84.5$\pm$4.7 &74.2$\pm$13.9 &76.4$\pm$8.1 &67.1$\pm$10.1 &80.1$\pm$9.6 &76.5 &27.2$\pm$24.7 &50.3$\pm$22.7 &45.7$\pm$23.5&49.6$\pm$30.9&29.8$\pm$20.1 &40.5\\
            BigAug~\cite{zhang2020generalizing} &84.2$\pm$5.0 &73.9$\pm$14.1 &73.3$\pm$7.7 &74.7$\pm$9.7 &79.0$\pm$6.8 &77.0 &35.9$\pm$26.3 &49.1$\pm$20.7 &53.8$\pm$22.0 &44.5$\pm$18.7 &28.9$\pm$14.4 &42.4\\
            JiGen~\cite{carlucci2019domain} &83.2$\pm$6.1 &70.8$\pm$14.7 &74.0$\pm$7.9 &71.5$\pm$10.2 &80.3$\pm$6.2 &75.9 &29.3$\pm$23.2 &64.5$\pm$23.3 &50.4$\pm$25.9 &50.6$\pm$26.0 &24.3$\pm$10.7 &43.8 \\

            \hline
            Baseline + ISD &85.4$\pm$5.1 &74.0$\pm$13.0 &79.0$\pm$4.4 &72.7$\pm$9.2 &80.8$\pm$10.0 &78.4 &19.4$\pm$21.3 &54.8$\pm$24.8 &44.2$\pm$21.7 &48.9$\pm$23.6 &34.9$\pm$26.4 &40.4\\ 
            Baseline + TTA &84.7$\pm$8.2 &75.2$\pm$8.2 &74.8$\pm$8.2 &67.2$\pm$8.2 &79.7$\pm$8.2 &76.3 &25.7$\pm$23.4 &42.0$\pm$20.2 &46.1$\pm$21.9 &47.9$\pm$37.4 &30.3$\pm$20.6 &38.4 \\
            TASD (\textit{Ours}) &\textbf{87.1$\pm$2.5} &\textbf{76.4$\pm$6.1} &\textbf{82.5$\pm$5.2} &\textbf{76.0$\pm$6.6} &\textbf{83.2$\pm$6.7} &\textbf{81.1} &19.3$\pm$21.3 &\textbf{39.1$\pm$17.5} &\textbf{38.7$\pm$12.2} &\textbf{43.4$\pm$14.2} &\textbf{21.0$\pm$17.5} &\textbf{32.3}\\

            \hline
            \hline

        \end{tabular}

    }}
\label{tab:comparisons_prostate}
\end{table*} 
\begin{table*}[!tbp]
    \renewcommand\arraystretch{1.2}
    \centering
        \caption{\small{Quantitative comparison of single domain generalization results on fundus image segmentation between different methods.}}
        \resizebox{0.78\textwidth}{!}{%
        \setlength\tabcolsep{5.0pt}
        \scalebox{1.00}{
        \begin{tabular}{c|c  c  c  c ||c  c c c }
    %       \hline
            \hline
            \hline
            Unseen Site & B &  C & D &Avg. & B &  C & D &Avg.\\
            \hline
            \hline
            &\multicolumn{4}{c||}{\textbf{Dice Coefficient (Dice,  mean$\pm$std)~$\uparrow$}} &\multicolumn{4}{c}{\textbf{Hausdorff Distance (HD,  mean$\pm$std)~$\downarrow$}}\\ 
            \hline
            \hline
            Baseline &83.2$\pm$8.2 &76.0$\pm$14.8 &88.8$\pm$5.7 &82.7 &27.4$\pm$15.3 &36.8$\pm$23.8 &21.5$\pm$10.3 &28.6\\
            \hline
            M-ADA~\cite{qiao2020learning} &85.9$\pm$9.3 &77.4$\pm$13.7 &90.6$\pm$4.8 &84.6 &22.7$\pm$11.2 &30.4$\pm$17.1 &13.8$\pm$7.5 &22.3\\
            TTT~\cite{sun2020test} &84.6$\pm$5.7 &77.3$\pm$11.9 &89.0$\pm$4.1 &83.7 &22.2$\pm$8.6 &31.4$\pm$19.1 &17.4$\pm$7.1 &23.7\\
            TTST~\cite{karani2020test} &85.5$\pm$9.0 &76.5$\pm$13.4 &91.0$\pm$4.7 &84.3 &22.0$\pm$10.2 &33.6$\pm$19.3 &\textbf{12.1$\pm$8.1} &22.6\\
            Tent~\cite{wang2020fully} &85.1$\pm$8.1&77.1$\pm$13.0&89.2$\pm$4.9&83.8 &23.1$\pm$11.7&35.2$\pm$20.9&17.8$\pm$8.5&25.4\\
            BigAug~\cite{zhang2020generalizing} &84.7$\pm$7.5 &78.0$\pm$13.5 &90.7$\pm$4.6 &84.5 &27.1$\pm$13.2 &30.3$\pm$21.6 &14.8$\pm$7.9 &24.1\\
            JiGen~\cite{carlucci2019domain} &84.5$\pm$5.8 &77.5$\pm$12.0 &88.5$\pm$4.1 &83.5 &23.4$\pm$9.2 &34.3$\pm$19.4 &20.4$\pm$10.6 &26.0\\
            
            \hline
            Baseline + ISD &85.7$\pm$8.0 &78.1$\pm$12.9 &89.5$\pm$5.9 &84.4 &22.7$\pm$10.0 &29.6$\pm$17.9 &15.2$\pm$9.2 &22.5\\ 
            Baseline + TTA &83.6$\pm$7.6 &77.2$\pm$13.8 &89.6$\pm$5.6 &83.5 &26.0$\pm$13.9 &34.1$\pm$22.6 &15.8$\pm$10.6 &25.3\\
            TASD (\textit{Ours}) &\textbf{87.6$\pm$8.0} &\textbf{78.5$\pm$12.6} &\textbf{91.3$\pm$4.2} &\textbf{85.8} &\textbf{19.8$\pm$9.5} &\textbf{29.4$\pm$18.0} &12.3$\pm$6.2 &\textbf{20.5}\\

            \hline
            \hline

        \end{tabular}

    }}
\label{tab:comparisons_fundus}
\end{table*}

\textbf{Comparison results:}
Table~\ref{tab:comparisons_prostate} shows the results on prostate MRI segmentation. We see that different methods can generally improve the generalization performance over baseline. The M-ADA approach dedicatedly designed for the SDG setting performs better than the DG methods with self supervision (JiGen) or traditional data augmentation (BigAug). Compared with M-ADA, our TASD presents clear Dice increase on all unseen sites, with improvement of 2.9\% and 8.7 on the average performance of Dice and HD respectively. This can be attributed to our explicit integration of shape priors and effective utilization of the prior knowledge at test time to improve generalization. For the test-time learning methods, TTST outperforms TTT and Tent in our segmentation problem, owing to their self-training strategy to gradually refine the segmentation masks. Our TASD performs clearly better than TTST in the average performance on both two metrics. This endorses the benefits of the general shape priors integrated to the task model, which provides strong reference for the test-time update to facilitate the segmentation at unseen domains. Overall, TASD achieves the best average performance and consistently outperforms the baseline on all unseen sites by a large margin.

\begin{figure}[t]
	\centering
	\includegraphics[width=0.48\textwidth]{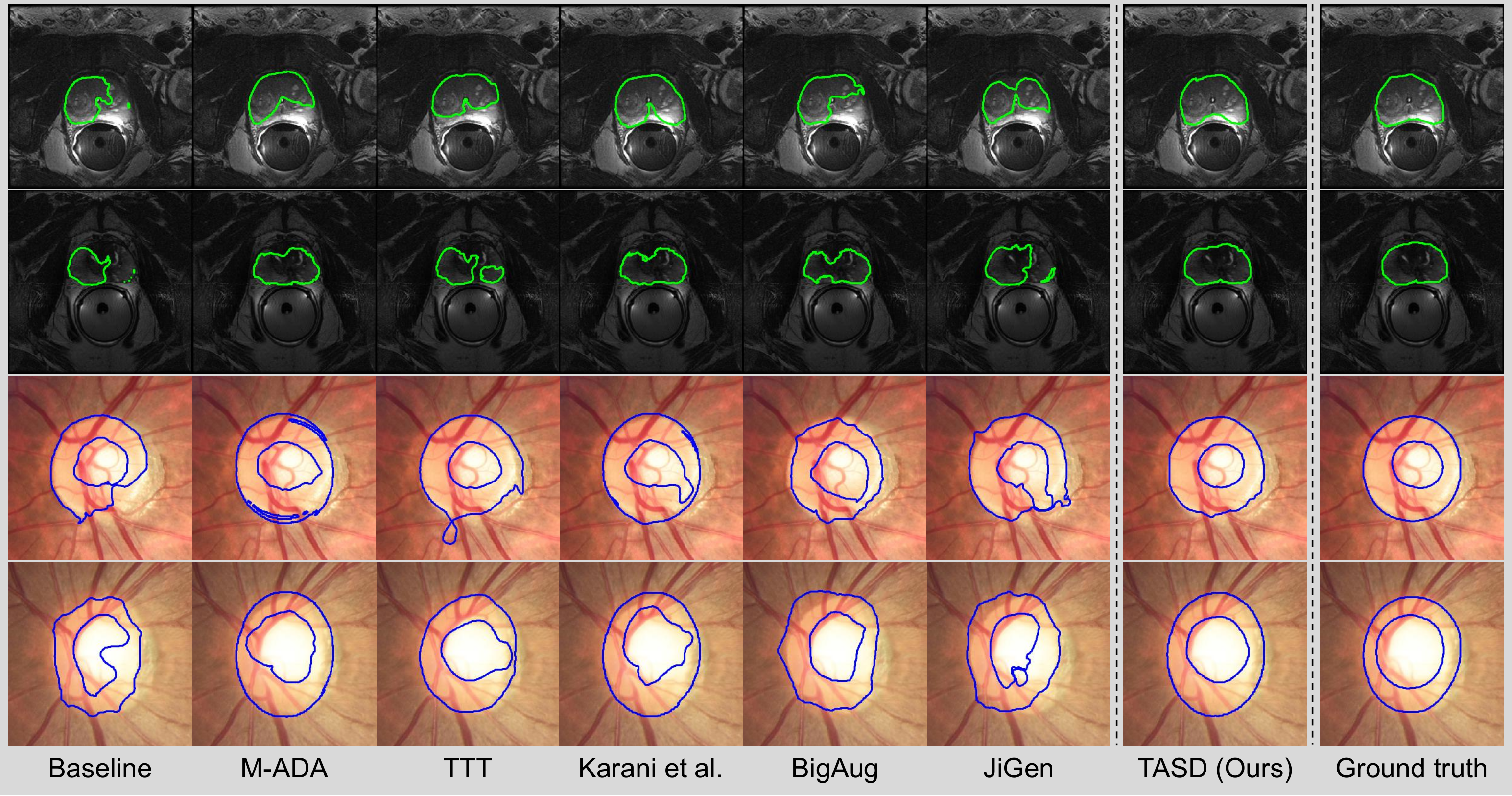}
    \caption{Qualitative comparison for different methods under SDG scenario, with first two rows for prostate MRI segmentation and last two rows for fundus image segmentation.}
 	\label{fig:qualitative_comparison}
\end{figure}

Table~\ref{tab:comparisons_fundus} shows the results on fundus image segmentation.
Similarly, all comparison methods can more or less improve the average generalization performance, and the other observations drawn in prostate MRI segmentation also hold here. Our TASD  beats all comparison methods in the average performance, with clear improvements over baseline on all three unseen sites. This again confirms the superiority of our method to address SDG for medical image segmentation.  Fig.~\ref{fig:qualitative_comparison} further shows the segmentation results with two example unseen cases for each task. It is observed that TASD can accurately preserve the anatomy shapes at unknown distributions, whereas other methods sometimes fail to do so.

\subsection{Ablation Study}
We conduct experiments to investigate several key points in our method: 1) the contribution of the two components of integration of shape dictionary (ISD) and test-time adaptation (TTA) in our method; 2) the representation power of the shape dictionary; 3) the effect of dictionary size and number of masks to learn the shape dictionary; 4) the effect of the two consistency regularization terms at test-time adaptation.

\textbf{Contribution of the two components.}
As shown in Table~\ref{tab:comparisons_prostate} and Table~\ref{tab:comparisons_fundus}, integrating the shape dictionary over baseline (Baseline + ISD) yields consistent improvements on different unseen sites, which demonstrates the benefits of utilizing the general shape information to enhance model generalization under the worst single domain training scenario. 
Further adapting the model at test time completes our TASD method, which increases the average Dice over Baseline + ISD by 2.7\% and 1.4\% for the two tasks. 
Notably, such improvements are higher than the performance gains when deploying test-time adaptation on the baseline model (i.e., Baseline + TTA vs. Baseline), which is 1.5\% and 0.8\% for the two tasks.
This observation reflect that our test-time adaptation mechanism gains additional capacity when deployed over the network with integrated shape dictionary, and enables the model to adaptively utilize the shape priors from single-domain data to improve model generalizability.

\begin{figure}[t]
	\centering
	\includegraphics[width=0.48\textwidth]{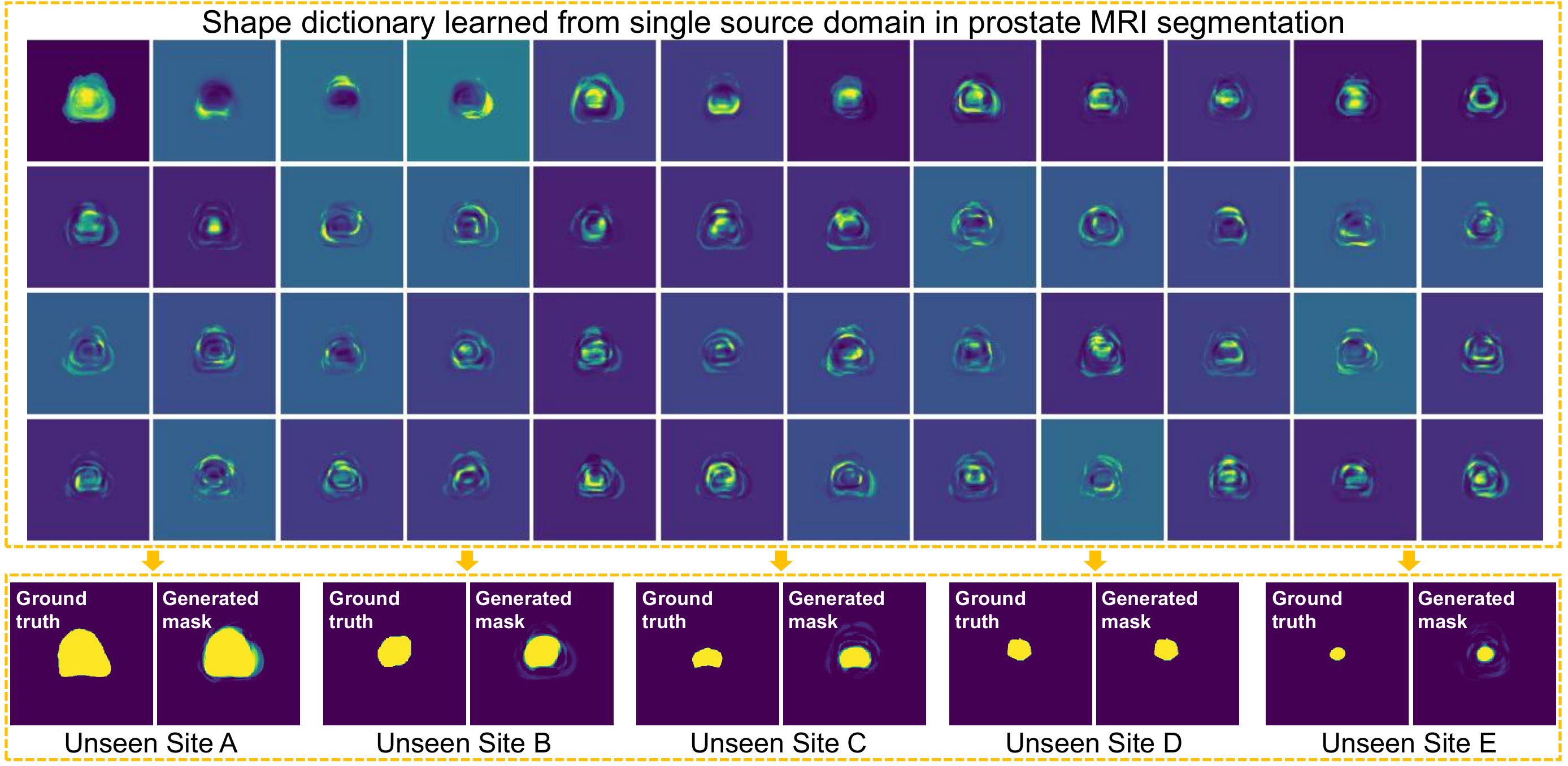}
    \caption{Visualization of the shape dictionary learned from single source domain with $K$=48, as well as the mask derived from the dictionary for a representative sample at each unseen site (with ground truth displayed for reference). }
 	\label{fig:shape_dictionary}
\end{figure}
\begin{figure}[t]
	\centering
	\includegraphics[width=0.48\textwidth]{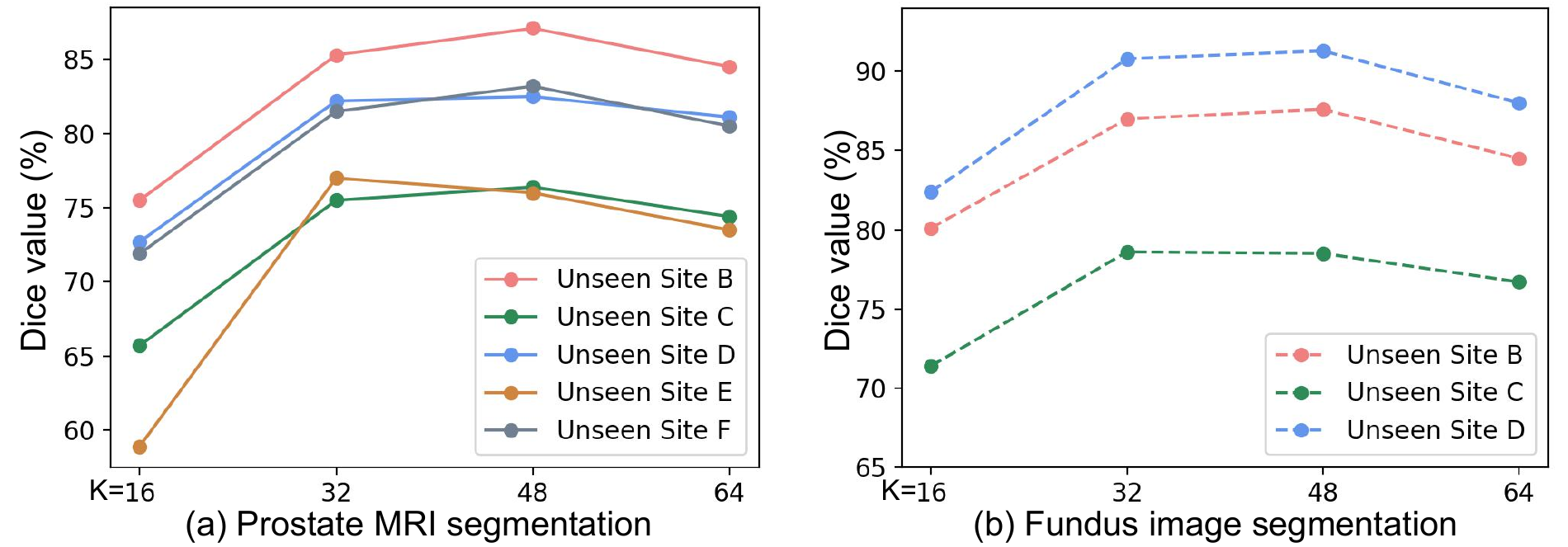}
    % \caption{Ablation analysis with different dictionary size $K$.} 
    \caption{Effect of different dictionary size $K$ on the generalization performance.}
 	\label{fig:ablation_dict_size}
\end{figure}

\textbf{Representation power of the learned shape dictionary.}
Fig.~\ref{fig:shape_dictionary} displays the learned shape dictionary from single-domain data with dictionary size $K$=48 and the mask derived from the dictionary for representative samples of each unseen site. We can see that the masks derived from pre-collected shape priors for different unseen-site samples are highly consistent with their ground truth masks with various sizes. We also directly compute the accuracy of reference mask $\bm{M}$ (cf. Eq.~\ref{eq:reference_mask}) generated from shape dictionary, observing that it attains 77.1\% and 83.9\% Dice for the averaged generalization performance in prostate and fundus tasks, which is clearly higher than the baseline model.
These results indicate that the shape dictionary learned from single-domain data is capable of representing the diverse masks at each unseen site, and affirm our motivation to utilize the general shape priors information to address the challenging single domain generalization problem.

\textbf{Effect of dictionary size $K$ and number of ground truth masks $N$.}
The choice of element number $K$ in dictionary is important in our method, which affects the representation power of the extracted shape priors. 
Intuitively, less elements in the shape dictionary (i.e., smaller $K$) might be incapable to represent the diverse segmentation masks, while too more elements (i.e., larger $K$) could lead to overfitting of the dictionary and make each shape template less representative to generalize to unseen domains.
To investigate the suitable choice of $K$, we repeat the experiment of TASD by varying $K$$\in$$\{16, 32, 48, 64\}$. As shown in Fig.~\ref{fig:ablation_dict_size}, the models with middle-level dictionary size ($K$=32 or 48) perform 
better than the model with smaller ($K$=16) or larger dictionary ($K$=64) on both tasks. These results confirm our analysis above, and we finally adopt $K$=48 in our method. Furthermore, we analyze the effect of $N$ which is the number of ground truth masks used to learn the shape dictionary. Specifically, we randomly sample a certain proportion from all masks of single-domain data, observing that the Dice performances decrease by 0.7\% and 1.6\% in prostate segmentation, 0.4\% and 0.9\% in fundus image segmentation tasks when sampling 50\% and 25\% from the total ground truth masks. This reflects that a larger $N$ is helpful to learn more general shape priors and therefore can better improve the generalization performance under data distribution shifts.

\begin{figure}[t]
	\centering
	\includegraphics[width=0.48\textwidth]{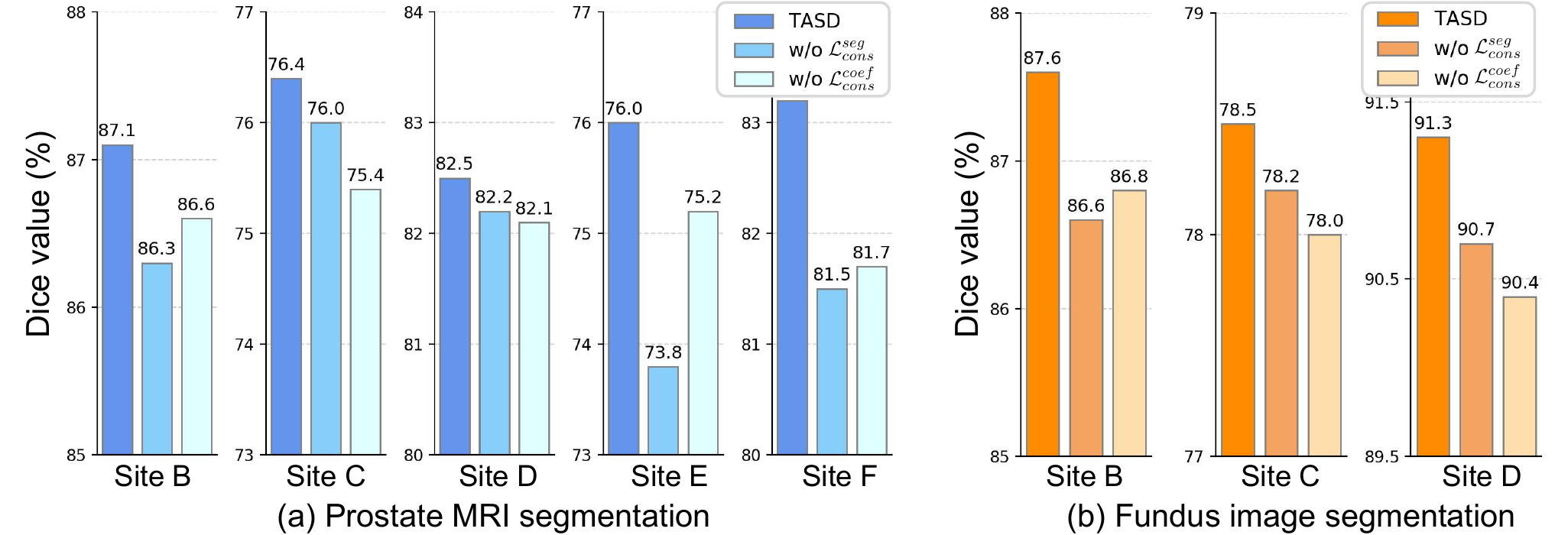}
    \caption{Ablation analysis of the dual-consistency regularization mechanism at test-time adaptation.} 
 	\label{fig:ablation_dict}
\end{figure}

\textbf{Effect of two consistency regularization terms.} To let the model adaptively exploit the integrated shape priors, we have designed a dual-consistency regularization mechanism which imposes the consistency of both segmentation and shape coefficient predictions under input perturbations. As listed in Fig.~\ref{fig:ablation_dict}, removing the regularization from either place (w/o $\mathcal{L}_{cons}^{coef}$ or w/o $\mathcal{L}_{cons}^{seg}$) can lead to performance drop compared with the full TASD model. This demonstrates the indispensable roles of the two regularization terms in our method. Specifically, the $\mathcal{L}_{cons}^{coef}$ allows the model to better utilize the shape prior knowledge at test time, while the $\mathcal{L}_{cons}^{seg}$ helps to better adapt the network parameters and refine the generated reference masks.

% \vspace{-1mm}

\section{Conclusion}
% \vspace{-1mm}
We present a new method with promising performance achieved on the challenging SDG problem in medical image segmentation. The idea is to integrate the semantic  shape priors that are generalizable across domains and can be well captured from single-domain data to assist segmentation, and further effectively exploit these prior information via test-time adaptation driven by dual-consistency regularization to promote the model generalization at any unseen data distributions. The integration of shape priors extracted with dictionary learning is extendable to address other segmentation problems such as multi-modality learning, and evaluating its benefit is appealing as our future work. Besides, the proposed test-time learning method with consistency regularization mechanism is applicable to other scenarios.

\section{Acknowledgement}
This work was supported by Key-Area Research and Development Program of Guangdong Province, China under Grant 2020B010165004, Hong Kong RGC TRS Project No. T42-409/18-R, Hong Kong Innovation and Technology Fund Project No. GHP/110/19SZ, and National Natural Science Foundation of China with Project No. U1813204.

% \bibliographystyle{aaai21}
% \input{reference.tex}

% \clearpage
\bibliography{refs}

\end{document}